%% file: main.tex
\documentclass[10pt,twocolumn,letterpaper]{article}
\usepackage{algorithm}
\usepackage{algorithmic}
\usepackage{booktabs}
\usepackage{graphicx}
\usepackage{longtable}
\usepackage{tcolorbox}
\usepackage{stackengine}
\usepackage[T1]{fontenc}
\usepackage[accsupp]{axessibility}
\usepackage[pagenumbers]{cvpr} 

\usepackage{enumitem}
\usepackage{marvosym}

\usepackage{array}
\usepackage{lscape}
\usepackage{caption}
\usepackage{makecell}

\usepackage{tabularx}

\usepackage{multirow}

\definecolor{cvprblue}{rgb}{0.21,0.49,0.74}
\usepackage[numbers,sort&compress]{natbib}
\usepackage[pagebackref,breaklinks,colorlinks=true,citecolor=cvprblue]{hyperref}

\title{FilterRAG: Zero-Shot Informed Retrieval-Augmented Generation to Mitigate Hallucinations in VQA}

\author{
Nobin Sarwar\textsuperscript{$\diamondsuit$} \qquad  \\
\textsuperscript{$\diamondsuit$}University of Maryland, Baltimore County\\
{\small \tt smsarwar96@gmail.com} 
}

\begin{document}
\maketitle
\input{sections/1_Introduction}
\input{sections/2_Background}
\input{sections/3_Method}

\input{sections/4_Experiment}
\input{sections/5_Conclusion}
{
    \small
    \bibliographystyle{ieeenat_fullname}
    \bibliography{main}
}

\appendix
\input{supplement.tex}
\end{document}

%% file: sections/1_Introduction.tex
\begin{abstract}
Visual Question Answering requires models to generate accurate answers by integrating visual and textual understanding. However, VQA models still struggle with hallucinations, producing convincing but incorrect answers, particularly in knowledge-driven and Out-of-Distribution scenarios. We introduce FilterRAG, a retrieval-augmented framework that combines BLIP-VQA with Retrieval-Augmented Generation to ground answers in external knowledge sources like Wikipedia and DBpedia. FilterRAG achieves 36.5\% accuracy on the OK-VQA dataset, demonstrating its effectiveness in reducing hallucinations and improving robustness in both in-domain and Out-of-Distribution settings. These findings highlight the potential of FilterRAG to improve Visual Question Answering systems for real-world deployment.
\end{abstract}

\section{Introduction}
\label{sec:introduction}

\begin{figure}[t]
    \centering
    \includegraphics[width=\linewidth]{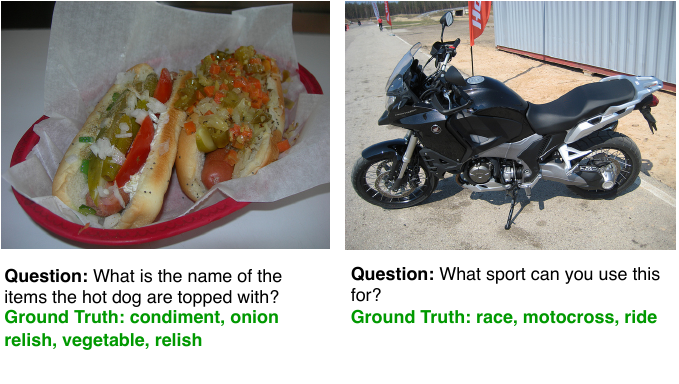}
    \caption{Two examples of question-answer pairs from the OK-VQA dataset. The left example asks about the items on a hot dog, requiring models to incorporate external knowledge of common food items. The right example asks about the sport associated with a motorcycle, emphasizing the need to understand how people typically use such vehicles. These examples illustrate the fundamental challenge of OK-VQA, where models rely on external knowledge to generate accurate answers rather than depending solely on the image.}
    \label{fig:ok-vqa_sample}
\end{figure}

In Visual Question Answering (VQA) system, models need to interpret images and provide accurate responses to natural language questions~\cite{antol2015vqa, tu2014joint, marino2019ok}. One major challenge in VQA is answering questions that require external knowledge beyond what is explicitly depicted in the image. Figure~\ref{fig:ok-vqa_sample} provides two examples from OK-VQA dataset, where recognizing hot dog toppings requires knowledge of condiments, and identifying the sport associated with a motorcycle requires understanding its common use. These examples highlight the importance of developing models that integrate visual perception with broader world knowledge to improve VQA performance.

Recent advancements in Vision-Language Models (VLMs), such as BLIP \cite{li2022blip} and CLIP \cite{radford2021learning}, have demonstrated significant progress by leveraging large-scale pretraining on multimodal datasets. However, these models often produce hallucinations, such as plausible but incorrect answers, when confronted with knowledge-intensive questions or Out-of-Distribution (OOD) inputs~\cite{jiang2024negative, zang2024overcoming, bordes2024introduction}. Hallucinations arise when models rely excessively on learned biases or lack access to relevant external knowledge~\cite{radford2021learning, jia2021scaling}.

\begin{figure*}[t]
    \centering
    \includegraphics[width=0.8\textwidth]{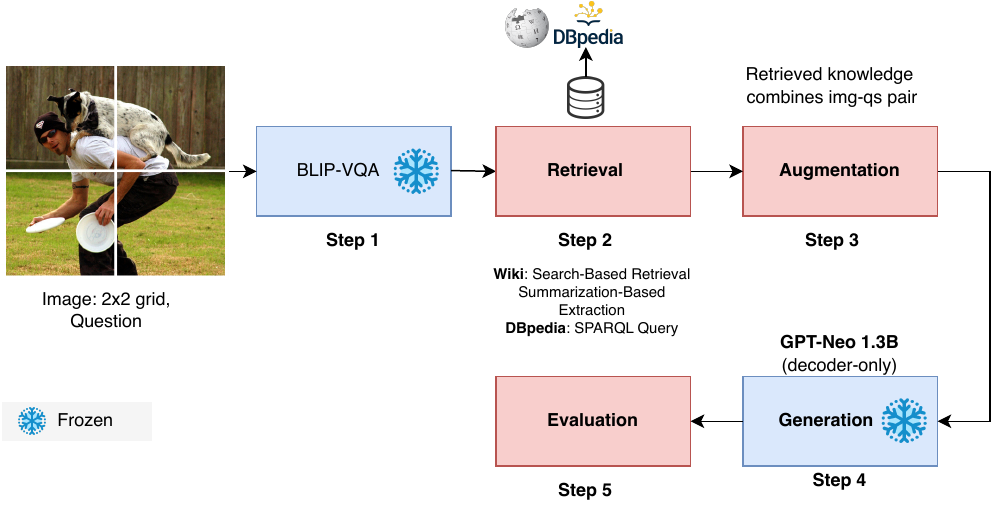}
    \caption{The \textbf{FilterRAG} architecture: A step-by-step process integrating frozen \textbf{BLIP-VQA} with \textbf{Retrieval-Augmented Generation (RAG)}. The system retrieves knowledge from Wikipedia and DBpedia, augments image-question pairs, and uses frozen \textbf{GPT-Neo 1.3B} to generate answers.}
    \label{fig:FilterRAG_Architecture}
\end{figure*}

To address these challenges, we introduce FilterRAG, a novel framework that integrates BLIP-VQA~\cite{li2022blip} with Retrieval-Augmented Generation (RAG)~\cite{lewis2020retrieval, ram2023context, karpukhin2020dense} to mitigate hallucinations in VQA, especially for OOD scenarios. FilterRAG grounds its answers in external knowledge sources such as Wikipedia and DBpedia, ensuring factually accurate and context-aware responses. The architecture, illustrated in Figure~\ref{fig:FilterRAG_Architecture}, employs a multi-step process: the input image is divided into a 2x2 grid to balance visual detail and coherence, visual and textual embeddings are generated using BLIP-VQA, and relevant knowledge is dynamically retrieved and integrated into the answer generation process using a frozen GPT-Neo 1.3B model~\cite{gpt-neo}.

In summary, we focus on three main challenges in Multimodal RAG based VQA:

\textbf{RQ1:} How can zero-shot learning improve retrieval and VQA accuracy to address hallucination in multimodal RAG systems?

\textbf{RQ2:} How does zero-shot learning contribute to better OOD performance in VQA models?

We evaluate FilterRAG on the OK-VQA dataset \cite{marino2019ok}, a benchmark requiring external knowledge beyond image content. Our results show that FilterRAG significantly reduces hallucinations compared to baseline models, achieving consistent performance across both in-domain and OOD settings. The qualitative analysis highlights the importance of effective knowledge retrieval and multimodal alignment for robust VQA.

To summarize, our key contributions are as follows:
\begin{itemize}
    \item \textbf{FilterRAG:} A retrieval-augmented approach that grounds VQA responses in external knowledge.
    
    \item \textbf{Zero-shot learning:} Enhancing retrieval and reducing hallucinations in OOD scenarios.
    
    \item \textbf{Comprehensive evaluation:} Evaluation on the OK-VQA dataset, demonstrating robustness and reliability for knowledge-intensive tasks.
\end{itemize}

This paper introduces FilterRAG, a retrieval-augmented framework to reduce hallucinations in VQA, especially in OOD scenarios. Section~\ref{sec:introduction} outlines the problem and motivation. Section~\ref{sec:Background} provides background knowledge on VLMs, VQA, RAG with VQA and OOD in VLMs. Section~\ref{sec:FilterRAG_Method} details the FilterRAG framework. Section~\ref{sec:Experiment} presents experiments on the OK-VQA dataset, including performance comparisons and ablation studies. Finally, Section~\ref{sec:Conclusion} summarizes findings and future directions.

%% file: sections/2_Background.tex
\section{Background}
\label{sec:Background}

\subsection{Vision Language Models}
Vision language models (VLMs) combine visual and linguistic data to understand and perform tasks requiring both image and text inputs~\cite{radford2021learning, jia2021scaling, singh2022flava}. By bridging the domains of Computer Vision (CV) and Natural Language Processing (NLP), these models can analyze complex scenes and respond meaningfully to textual descriptions, instructions, or queries. VLMs use multimodal embeddings to represent images and text in a shared feature space. This shared representation allows VLMs to align visual and textual information, supporting tasks like pairing images with captions or locating specific objects in images based on textual instructions. FilterRAG adopts the BLIP framework, leveraging its Multimodal mixture of Encoder-Decoder (MED) architecture for consistent visual and textual data processing in VQA~\cite{li2022blip}. This unified approach reduces memory usage and training time by sharing parameters between the encoder and decoder. As a result, BLIP enables faster inference without compromising accuracy, making it ideal for deployment in resource-constrained environments.

VLMs enable advanced applications such as VQA~\cite{antol2015vqa, zhang2016yin, goyal2017making}, image-text retrieval~\cite{luo2023end}, and image captioning~\cite{li2024lamp, covert2024locality}, expanding human-computer interaction capabilities. Despite these advancements, cross-modal alignment poses ongoing challenges, as aligning visual and linguistic data involves resolving complex ambiguities. Therefore, to ensure the safe and ethical deployment of VQA systems, we propose FilterRAG, a multimodal RAG framework. FilterRAG addresses hallucinations by grounding responses in retrieved external knowledge, enhancing robustness in OOD scenarios. By integrating multimodal retrieval with generative reasoning, our proposed approach effectively generalizes beyond the training knowledge base, providing accurate and context-aware answers to VQA queries.

\subsection{Visual Question Answering}
Visual Question Answering (VQA)~\cite{antol2015vqa, marino2019ok, zhang2016yin, goyal2017making} is a multimodal task that combines computer vision for image analysis (I) with natural language processing for question comprehension (Q) to generate accurate answers (A) about visual content. Recent VQA models, such as ViLBERT~\cite{lu2019vilbert}, VisualBERT~\cite{li2019visualbert}, VL-BERT~\cite{su2019vl}, and LXMERT~\cite{tan2019lxmert}, have significantly progressed through large-scale vision-language pretraining and sophisticated attention mechanisms. Their pretraining on large, diverse datasets, such as VQA 2.0~\cite{goyal2017making}, OK-VQA~\cite{marino2019ok}, VizWiz~\cite{bigham2010vizwiz}, and TDIUC~\cite{kafle2017analysis}, enables them to generalize well across various VQA tasks, improving performance on benchmarks requiring complex reasoning, multi-object interactions, and contextual understanding. Despite their advancements, these models frequently produce hallucinations and fail in OOD settings, a consequence of biased pretraining data that limits their robustness and adaptability.

To address these limitations, we propose a robust VQA framework that integrates BLIP-VQA~\cite{li2022blip} with RAG. By retrieving external knowledge, RAG grounds answers in factual information and improves performance on OOD queries. This retrieval mechanism expands the model knowledge beyond the training data, enhancing robustness and generalization. Our approach demonstrates significant improvements in answer accuracy on benchmarks such as VQA 2.0~\cite{goyal2017making} and OK-VQA~\cite{marino2019ok}. By unifying the BLIP architecture with retrieval-augmented techniques, the framework generates context-aware and reliable answers, making it suitable for real-world, dynamic environments.

\subsection{Retrieval-Augmented Generation with VQA}
Retrieval-Augmented Generation (RAG) enhances the effectiveness of VLMs by integrating external knowledge dynamically~\cite{lewis2020retrieval, ram2023context, karpukhin2020dense}. When a query involving visual and textual inputs is provided, the retriever searches external databases (e.g., Wikipedia) for relevant information. This retrieved content supplements the query, providing richer context. The generator then conditions its output on both the retrieved knowledge and the original query, producing more accurate, contextually grounded, and factually consistent responses~\cite{guo2022survey}. RAG, combined with VQA, effectively demonstrates significant progress in overcoming issues like hallucinations and poor OOD generalization. Recent works such as KAT~\cite{gui2021kat}, MAVEx~\cite{wu2022multi}, KRISP~\cite{marino2021krisp}, ConceptBERT~\cite{garderes2020conceptbert}, and EnFoRe~\cite{wu2022entity} focus on integrating external knowledge sources like Wikidata, Wikipedia, ConceptNet, or even web-based sources like Google Images~\cite{wu2022multi} to improve VQA systems. These methods use different strategies to fuse external knowledge with image and question inputs, whether by retrieving facts, aggregating knowledge graph nodes, or augmenting transformer-based architectures.

Despite advancements in methods like KAT~\cite{gui2021kat}, MAVEx~\cite{wu2022multi}, KRISP~\cite{marino2021krisp}, and ConceptBERT~\cite{garderes2020conceptbert}, these approaches often rely on external knowledge sources that may lack coverage for OOD  scenarios. Techniques such as RASO~\cite{fu2023generate} and TRiG~\cite{gao2022thousand} mitigate biases through answer refinement but struggle with noisy or irrelevant retrievals. Region-based methods like REVIVE~\cite{lin2022revive} and Mucko~\cite{zhu2020mucko} face scalability issues due to high-resolution processing demands. FilterRAG addresses these challenges by combining RAG with VLMs to enhance VQA performance in OOD settings, reducing hallucinations through efficient, contextually relevant retrieval. This approach improves upon existing works while maintaining computational efficiency, particularly for datasets like OK-VQA.

\subsection{Out-of-Distribution Detection in VLMs}
Out-of-Distribution (OOD) detection enhances model robustness by recognizing inputs that fall outside the training data distribution. Early work, such as~\cite{hendrycks2016baseline}, introduces a simple and effective method for OOD detection using the maximum softmax probability as a confidence score, where lower confidence scores indicate potential OOD data or misclassified inputs. In VLMs, OOD detection becomes more challenging due to multimodal representation shifts that occur when the model encounters novel or unseen data combinations. These shifts impact both the visual and textual data and, more importantly, how the two modalities interact within the latent space~\cite{tan2019lxmert, lu2019vilbert}.

For VLMs, given an input pair \( (x_v, x_t) \), where \( x_v \) is a visual input and \( x_t \) is a textual input, the task is to detect whether either the visual, textual, or their combined representation is OOD~\cite{dong2024multiood, jiang2024negative, duong2023general, zang2024overcoming, bordes2024introduction}. The embeddings from the two modalities, \( z_v = g_v(x_v) \) and \( z_t = g_t(x_t) \), are fused in a joint embedding space. The prediction probability \( \hat{p} \) can be obtained by a classifier \( h(\cdot) \) applied on the fused embeddings:

\begin{equation}
\hat{p} = \delta(h([z_v, z_t])) = \delta(h([g_v(x_v), g_t(x_t)])),
\end{equation}

where \( \delta(\cdot) \) is the softmax function, and \( h(\cdot) \) is a classifier.

In some methods, each modality can be checked for OOD status independently using separate classifiers \( h_v \) and \( h_t \) for vision and text:

\begin{equation}
\hat{p}_v = \delta(h_v(g_v(x_v))), \quad \hat{p}_t = \delta(h_t(g_t(x_t))),
\end{equation}

Finally, a threshold-based decision rule can be applied to classify the input as either In-Distribution (ID) or Out-of-Distribution (OOD). If the score \( S(x_v, x_t) \) exceeds a certain threshold \( \lambda \), the input is considered ID; otherwise, it is classified as OOD:

\begin{equation}
G_\lambda(x_v, x_t) = 
    \begin{cases} 
       \text{ID}, & \text{if } S(x_v, x_t) \geq \lambda \\
       \text{OOD}, & \text{if } S(x_v, x_t) < \lambda 
    \end{cases}
\end{equation}

%% file: sections/3_Method.tex
\section{The FilterRAG Method}
\label{sec:FilterRAG_Method}

\subsection{Overview}
FilterRAG integrates BLIP-VQA~\cite{li2022blip} with RAG to mitigate hallucinations in VQA, particularly in OOD scenarios. The architecture, illustrated in Figure~\ref{fig:FilterRAG_Architecture}, employs a multi-step process to ground VQA responses in external knowledge sources such as Wikipedia and DBpedia. The process begins by dividing the input image into a 2x2 grid to capture critical visual features while minimizing fragmentation. BLIP-VQA generates multimodal embeddings by encoding both the image and the associated question. The retrieval component then queries external knowledge sources, such as Wikipedia (using search-based and summarization techniques) and DBpedia (via SPARQL queries), to fetch relevant contextual information.

This retrieved knowledge is combined with the image-question pair, enriching the context for answer generation. A frozen GPT-Neo 1.3B~\cite{gpt-neo} model leverages this augmented information to produce the final answer. By grounding responses in retrieved factual data, FilterRAG effectively reduces hallucinations and enhances robustness, particularly for knowledge-intensive and OOD queries. Through the integration of external knowledge and efficient multimodal alignment, FilterRAG significantly improves the reliability and generalization of VQA systems, making it suitable for deployment in real-world applications where unseen concepts are common.

\subsection{Zero-Shot Learning in RAG Setting}
Zero-Shot Learning (ZSL)~\cite{xian2018zero, wang2019survey} enables models to generalize to unseen tasks or domains without requiring task-specific training data. For the VQA context, ZSL involves providing a model with an image $(I)$ and a question $(Q)$ and expecting it to produce accurate answers $(A)$ without fine-tuning task-specific datasets. Recent advancements in VLMs such as CLIP~\cite{radford2021learning}, ALIGN~\cite{jia2021scaling}, Frozen~\cite{tsimpoukelli2021multimodal}, and Flamingo~\cite{alayrac2022flamingo} have demonstrated robust performance across multiple downstream tasks through large-scale pretraining and multimodal alignment. Language Models (LMs) have also proven effective for Zero-Shot Learning through models like GPT-3~\cite{brown2020language} and T0~\cite{sanh2021multitask}, which leverage large-scale textual pretraining to perform a wide range of tasks without task-specific fine-tuning. 

Our method leverages BLIP-VQA~\cite{li2022blip} and the decoder-only language model GPT-Neo 1.3B~\cite{gpt-neo} within a Zero-Shot Learning setting. BLIP-VQA first aligns visual and textual features using its MED architecture. GPT-Neo 1.3B then utilizes this aligned context, along with the image description and question, to generate coherent and contextually relevant answers. To enhance robustness to OOD queries and reduce hallucinations, FilterRAG incorporates RAG, dynamically grounding responses in external knowledge sources. Our approach demonstrates strong performance on benchmarks like OK-VQA~\cite{marino2019ok}, which require knowledge beyond visual content.

\subsection{Visual Question Answering in Ok-VQA}
In the Visual Question Answering (VQA) task~\cite{antol2015vqa, marino2019ok, zhang2016yin, goyal2017making}, the goal is to predict the most appropriate answer ($A$) to a given question ($Q$) about an image ($I$). This relationship can be mathematically formalized as:
\begin{equation}
\hat{A} = \arg\max_{A \in \mathcal{A}} P(A \mid I, Q)
\end{equation}

where $A$ represents a possible answer, $I$ corresponds to the input image, and $Q$ denotes the input question. The OK-VQA dataset~\cite{marino2019ok} focuses specifically on open-domain questions that require external knowledge beyond the visual content of the image. Therefore, effective models for OK-VQA must combine visual and textual understanding with the ability to retrieve relevant external knowledge, ensuring accurate and context-aware responses.

VLMs generate the answer ($A$) as an open-ended sequence (e.g., free text), conditioned on both the image ($I$) and question ($Q$)~\cite{li2024configure}.  This can be formalized as:
\begin{equation}
P(\hat{A}) = \prod_{t=1}^{T} P(a_t \mid a_{1:t-1}, I, Q)
\end{equation}

where \( a_t \) denotes the token at time step \( t \) and \( a_{1:t-1} \) represents the preceding tokens. 

\subsection{Problem Formulation for RAG with VQA}
The objective of integrating RAG~\cite{lewis2020retrieval, ram2023context, karpukhin2020dense} with VQA is to predict the most accurate answer $A$ to a given question $Q$ about an image $I$ by leveraging both visual content and external knowledge retrieval. This process can be expressed probabilistically as:

\begingroup
\scriptsize
\begin{equation}
P_{\text{RAG}}(\hat{A}) \approx \prod_{i} \sum_{z \in \text{top-k}(p_\eta(\cdot \mid I, Q))} p_\eta(z \mid I, Q) p_\theta(a_i \mid I, Q, z, a_{1:i-1})
\end{equation}
\endgroup
\normalsize

Where \( z \) represents retrieved knowledge from an external corpus, \( p_\eta(z \mid I, Q) \) is the probability of retrieving \( z \) based on the image \( I \) and the question \( Q \), and \( p_\theta(a_i \mid I, Q, z, a_{1:i-1}) \) models the likelihood of generating the \( i \)-th token of the answer \( A \), conditioned on the previous tokens \( a_{1:i-1} \). In this formulation, the retriever \( p_\eta \) aims to fetch relevant knowledge \( z \) by leveraging both the visual content and the textual query. The retrieval process can be described as:
\begin{equation}
p_\eta(z \mid I, Q) \propto \exp\left(\mathbf{d}(z)^\top \mathbf{q}(I, Q)\right),
\end{equation}

where \( \mathbf{d}(z) \) is the embedding of the retrieved knowledge \( z \), and \( \mathbf{q}(I, Q) \) is the joint embedding of the image and the question. This formulation leverages a dual-encoder framework, similar to dense passage retrieval techniques~\cite{karpukhin2020dense}, and is further influenced by models such as Fusion-in-Decoder (FiD)~\cite{izacard2020leveraging}.

\subsection{OOD detection in VQA}
In Visual Question Answering (VQA), given an image \( I \) and a question \( Q \), the objective of out-of-distribution (OOD) detection is to determine whether the input pair belongs to the in-distribution dataset \( D_{\text{in}} \) or an OOD dataset \( D_{\text{OOD}} \)\cite{dong2024multiood, jiang2024negative, duong2023general, zang2024overcoming, bordes2024introduction}. This can be achieved using a scoring function \( S(I, Q) \) and a threshold \( \lambda \). The decision rule is defined as:
\begin{equation}
\footnotesize	
(I, Q) \in D_{\text{in}} \quad \text{if} \quad S(I, Q) \geq \lambda, \quad \text{else} \quad (I, Q) \in D_{\text{OOD}}.
\end{equation}

where \(D_{\text{in}}\) refers to the in-distribution dataset, \(D_{\text{OOD}}\) denotes the out-of-distribution dataset, \(S(I, Q)\) is the scoring function that computes the confidence for the pair, and \(\lambda\) is the threshold for distinguishing between \(D_{\text{in}}\) and \(D_{\text{OOD}}\). 

Our approach integrates these techniques within a RAG framework. By combining retrieval confidence with generation confidence, our scoring function $S(I,Q)$ captures both visual and knowledge-based uncertainties. This hybrid strategy improves OOD detection, enabling the model to flag uncertain inputs and enhancing the robustness of VQA systems.

\subsection{Binary Cross-Entropy Loss}
Binary cross-entropy loss is a standard measure for evaluating the correctness of predictions in classification tasks, including VQA. It is formulated as:
\begin{equation}
\mathcal{L} = - \frac{1}{n} \sum_{i=1}^n \left[ y_i \cdot \log(p_i) + (1 - y_i) \cdot \log(1 - p_i) \right]
\end{equation}

where \( n \) is the total number of predictions, \( y_i \) represents the ground-truth label for the \( i \)-th sample (\( y_i \in \{0, 1\} \)), and \( p_i \) is the predicted probability that the \( i \)-th sample belongs to the positive class.

In VQA, where answers can be evaluated against multiple valid responses, this loss function helps optimize model performance by reducing uncertainty and improving prediction accuracy~\cite{antol2015vqa, goyal2017making}. Models such as ViLBERT~\cite{lu2019vilbert} and LXMERT~\cite{tan2019lxmert} have effectively utilized binary cross-entropy loss to enhance their training processes, ensuring more reliable and accurate VQA outputs.

\subsection{Hallucination}
Grounding score \( g_{\text{mean}}(\hat{A}) \) quantifies semantic alignment between a predicted answer \(\hat{A}\) and ground truth answers in VQA. Using cosine similarity~\cite{radford2021learning, jia2021scaling} , the grounding score is:
\begin{equation}
g_{\text{mean}}(\hat{A}) = \frac{1}{n} \sum_{i=1}^n \frac{\mathbf{v}_{\text{pred}} \cdot \mathbf{v}_{\text{gt}}^i}{\|\mathbf{v}_{\text{pred}}\| \|\mathbf{v}_{\text{gt}}^i\|}
\end{equation}

where \( n \) is the number of ground truth answers, \( \mathbf{v}_{\text{pred}} \) is the embedding of the predicted answer \( \hat{A} \), and \( \mathbf{v}_{\text{gt}}^i \) is the embedding of the \( i \)-th ground truth answer. This grounding score measures the degree of alignment between the predicted and ground truth answers, capturing semantic similarity even when the answers differ lexically. Embedding models like word2vec~\cite{mikolov2013efficient}, GloVe~\cite{pennington2014glove}, and contextual models such as BERT~\cite{devlin2018bert} are commonly used to generate these embeddings. However, our approach replaces these traditional models with the more efficient Sentence Transformers (all-MiniLM-L6-v2)~\cite{reimers2019sentence}. This model produces compact and high-quality embeddings, enabling accurate measurement of alignment between predicted and ground truth answers while maintaining computational efficiency.   

Hallucination~\cite{zhou2020detecting, maynez2020faithfulness} is detected when the grounding score falls below a predefined threshold \( \tau \), indicating a lack of semantic alignment between the predicted answer and the ground truth:
\begin{equation}
\text{Hallucination} \quad \text{if} \quad g_{\text{mean}}(\hat{A}) < \tau
\end{equation}

Hallucinations occur when models generate plausible yet incorrect answers that are not supported by the input context. However, this problem is common in models like CLIP~\cite{radford2021learning} and BLIP~\cite{li2022blip} due to the reliance on learned biases. To address this challenge, our approach integrates BLIP-VQA~\cite{li2022blip} with RAG for fact-grounded answers. We enhance robustness by incorporating OOD detection to identify queries beyond the training data and applying a grounding score to measure semantic alignment. This combined strategy effectively reduces hallucinations and ensures accurate, context-aware answers.

%% file: sections/4_Experiment.tex
\section{Experiment}
\label{sec:Experiment}

\subsection{Dataset}
Outside Knowledge Visual Question Answering (OK-VQA)~\cite{marino2019ok} is a benchmark dataset designed to evaluate VQA systems that require leveraging external knowledge sources beyond the information present in an image. The dataset consists of 14,055 knowledge-based questions paired with 14,031 images from the COCO dataset~\cite{lin2014microsoft}. These questions span 10 diverse knowledge categories, including domains such as Science and Technology, Geography, Cooking and Food, and Vehicles and Transportation. The questions were crowdsourced via Amazon Mechanical Turk, ensuring they require real-world knowledge to answer, making this dataset significantly more challenging than conventional VQA datasets. 

The dataset is split into 9,009 training samples and 5,046 testing samples, with each question associated with 10 ground-truth answers annotated by human annotators. This multi-answer format helps address ambiguity and variability in responses. Table~\ref{tab:okvqa_details} outlines key statistics and the distribution of questions across various knowledge categories in the Ok-VQA dataset. Baseline evaluations on OK-VQA using state-of-the-art models like MUTAN and Bilinear Attention Networks (BAN) reveal a significant drop in performance compared to traditional VQA datasets. This performance degradation underscores the need for models with enhanced retrieval and reasoning capabilities to incorporate unstructured, open-domain knowledge effectively.

\begin{table}[h!]
    \centering
    \footnotesize
    \setlength{\tabcolsep}{4pt}
    \renewcommand{\arraystretch}{1.2}
    \caption{Key Details of the OK-VQA Dataset}
    \begin{tabular}{|p{3.2cm}|p{4.8cm}|}
        \hline
        \textbf{Attribute}                & \textbf{Details} \\
        \hline
        \textbf{Name}                     & OK-VQA (Outside Knowledge VQA) \\
        \hline
        \textbf{Source}                   & COCO Image Dataset \\
        \hline
        \textbf{Number of Questions}      & 14,055 \\
        \hline
        \textbf{Number of Images}         & 14,031 \\
        \hline
        \textbf{Question Categories}      & 10 Categories \\
        \hline
        \textbf{Categories Breakdown}     & Vehicles \& Transportation (16\%) \newline Brands, Companies \& Products (3\%) \newline Objects, Materials \& Clothing (8\%) \newline Sports \& Recreation (12\%) \newline Cooking \& Food (15\%) \newline Geography, History, Language \& Culture (3\%) \newline People \& Everyday Life (9\%) \newline Plants \& Animals (17\%) \newline Science \& Technology (2\%) \newline Weather \& Climate (3\%) \newline Other (12\%) \\
        \hline
        \textbf{Average Question Length}  & 8.1 words \\
        \hline
        \textbf{Average Answer Length}    & 1.3 words \\
        \hline
        \textbf{Unique Questions}         & 12,591 \\
        \hline
        \textbf{Unique Answers}           & 14,454 \\
        \hline
        \textbf{Answer Annotations}       & 10 answers per question \\
        \hline
        \textbf{Answer Types}             & Open-ended \\
        \hline
        \textbf{Requires External Knowledge} & Yes (e.g., Wikipedia, Common Sense, etc.) \\
        \hline
        \textbf{Typical Knowledge Sources}& Unstructured Text (Wikipedia) \\
        \hline
    \end{tabular}
    \label{tab:okvqa_details}
\end{table}

\subsection{Implementation Details}
The experiments are conducted on Google Colab using a T4 GPU. The NVIDIA T4 GPU features 16 GB of GDDR6 memory, 320 Tensor Cores, and supports mixed-precision computation, making it suitable for deep learning tasks. Due to computational constraints, we evaluate our model on a subset of 100 samples from the OK-VQA dataset~\cite{marino2019ok}.

\subsection{OOD and ID Category Splits}
In our experiments, we evaluate our approach using the OK-VQA dataset~\cite{marino2019ok}, which we split into OOD and ID subsets based on knowledge categories. The OOD categories include Vehicles and Transportation, Brands, Companies and Products, Sports and Recreation, Science and Technology, and Weather and Climate. The ID categories comprise Objects, Materials and Clothing, Cooking and Food, Geography, History, Language and Culture, People and Everyday Life, Plants and Animals, and Other. Using this split, we can assess how well the model generalizes to different categories of knowledge.

\subsection{Patch-Based Image Preprocessing}
For VQA processing, we preprocess each input image by dividing it into patches of various sizes, specifically 2×2, 3×3 and 4x4 grids. This patch-based approach captures fine-grained visual details, which can enhance feature extraction for complex queries. We then employ the BLIP-VQA model~\cite{li2022blip} to extract image representations and generate initial contextual information based on the image and the associated question.

\subsection{Retrieval-Augmented Knowledge Integration}
To incorporate external knowledge, we use  RAG~\cite{lewis2020retrieval} with external knowledge sources such as Wikipedia and DBpedia. RAG retrieves relevant information based on the question and the visual features extracted by BLIP-VQA~\cite{li2022blip}. This retrieval process supplies the model with real-world context beyond the image, which is crucial for correctly answering questions that depend on external knowledge.

\subsection{State-of-the-Art Performance Comparison}
We evaluate our proposed FilterRAG framework on the OK-VQA dataset and compare it to state-of-the-art methods (Table~\ref{table:SOTA-OK-VQA}). The baseline models, Base1 and Base2, use the BLIP-VQA model with the VQA v2~\cite{goyal2017making} and OK-VQA datasets~\cite{marino2019ok}, achieving 83.0\% and 40.0\% accuracy, respectively. The drop highlights the challenge of knowledge-based questions in OK-VQA. Our FilterRAG framework, which integrates BLIP-VQA, RAG, and external knowledge sources like Wikipedia and DBpedia, achieves 36.5\% accuracy in OOD settings. This result demonstrates the effectiveness of grounding VQA responses with external knowledge, especially for OOD scenarios. 

Compared to state-of-the-art methods, KRISP~\cite{marino2021krisp}  achieves 38.35\% with Wikipedia and ConceptNet, while MAVEx~\cite{wu2022multi} reaches 41.37\% using Wikipedia, ConceptNet, and Google Images. The highest performance comes from KAT (ensemble)~\cite{gui2021kat} at 54.41\% with Wikipedia and Frozen GPT-3. Although these models achieve higher accuracy, they often require significant computational resources. 

\begin{figure}[h!]
    \centering
    \includegraphics[width=\linewidth]{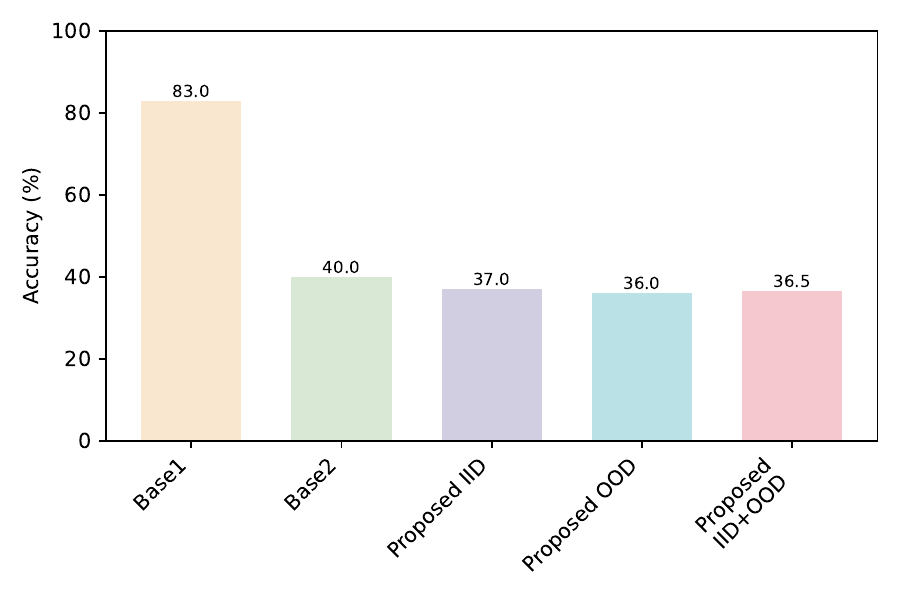}
    \caption{Comparison of Model Accuracy Across Different Settings.}
    \label{fig:plot1_accuracy}
\end{figure}

FilterRAG balances performance and efficiency, making it suitable for resource-constrained environments. As shown in Figure~\ref{fig:plot1_accuracy}, it achieves 37.0\% accuracy in ID settings, 36.0\% in OOD settings, and 36.5\% when combining ID and OOD data. This highlights its robustness for knowledge-intensive VQA tasks.

\begin{table*}[t]
    \centering
    \footnotesize
    \caption{Performance Comparison of State-of-the-Art Methods on the OK-VQA Dataset}
    \label{tab:okvqa_results}
    \renewcommand{\arraystretch}{1.2}
    \setlength{\tabcolsep}{10pt}
    \begin{tabular}{l l c}
        \toprule
        \textbf{Method}                                & \textbf{External Knowledge Sources}                          & \textbf{Accuracy (\%)} \\
        \midrule
        Q-only (Marino et al., 2019)~\cite{marino2019ok}                  & —                                                          & 14.93                  \\
        MLP (Marino et al., 2019)~\cite{marino2019ok}                     & —                                                          & 20.67                  \\
        BAN (Marino et al., 2019)~\cite{marino2019ok}              & —                                                          & 25.1                  \\
        MUTAN (Marino et al., 2019)~\cite{marino2019ok}               & —                                                          & 26.41                  \\
        ClipCap (Mokady et al., 2021)~\cite{mokady2021clipcap}                 & —                                                          & 22.8                   \\
        \midrule
        BAN + AN (Marino et al., 2019~\cite{marino2019ok}                  & Wikipedia                                                  & 25.61                  \\
        BAN + KG-AUG (Li et al., 2020)~\cite{li2020boosting}        & Wikipedia + ConceptNet                                     & 26.71                  \\
        Mucko (Zhu et al., 2020)~\cite{zhu2020mucko}                      & Dense Caption                                              & 29.2                   \\
        ConceptBERT (Gardères et al., 2020)~\cite{garderes2020conceptbert}           & ConceptNet                                                 & 33.66                  \\
        KRISP (Marino et al., 2021)~\cite{marino2021krisp}                   & Wikipedia + ConceptNet                                     & 38.35                  \\
        RVL (Shevchenko et al., 2021)~\cite{shevchenko2021reasoning}                 & Wikipedia + ConceptNet                                     & 39.0                   \\
        Vis-DPR (Luo et al., 2021)~\cite{luo2021weakly}                    & Google Search                                              & 39.2                   \\
        MAVEx (Wu et al., 2022)~\cite{wu2022multi}                       & Wikipedia + ConceptNet + Google Images                    & 41.37                  \\
        PICa-Full (Yang et al., 2022)~\cite{yang2022empirical}                 & Frozen GPT-3 (175B)                                        & 48.0                   \\
        KAT (Gui et al., 2022) (Ensemble)~\cite{gui2021kat}             & Wikipedia + Frozen GPT-3 (175B)                           & 54.41                  \\
        REVIVE (Lin et al., 2022) (Ensemble)~\cite{lin2022revive}          & Wikipedia + Frozen GPT-3 (175B)                           & 58.0                   \\
        RASO (Fu et al., 2023)~\cite{fu2023generate}                        & Wikipedia + Frozen Codex                                   & 58.5                   \\
        \midrule
        \textbf{FilterRAG (Ours)}                     & Wikipedia + DBpedia (\textbf{Frozen} BLIP-VQA and GPT-Neo 1.3B)    & \textbf{36.5}          \\
        \bottomrule
    \end{tabular}
    \label{table:SOTA-OK-VQA}
\end{table*}

\subsection{Hallucination Detection via Grounding Scores}
\begin{figure}[h!]
    \centering
    \includegraphics[width=\linewidth]{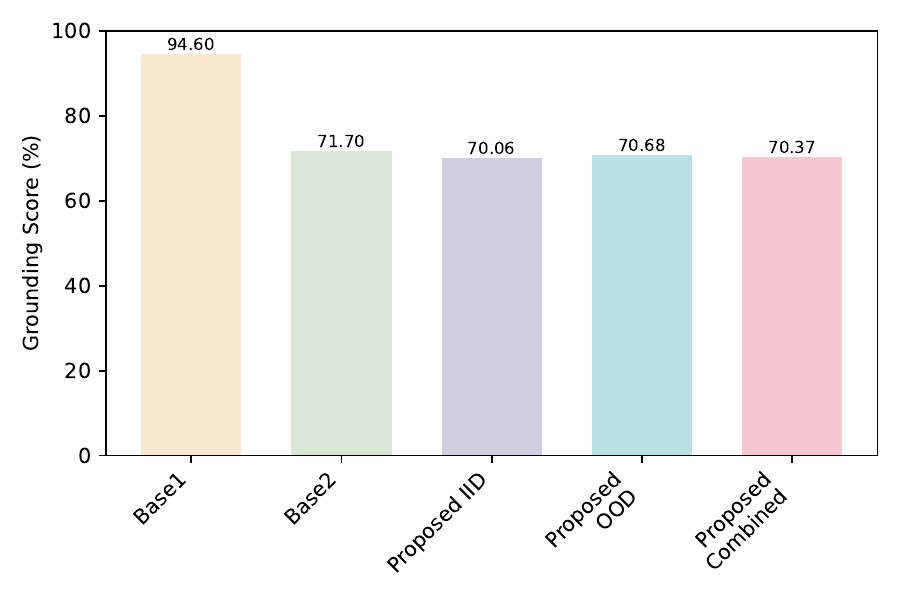}
    \caption{Grounding Score Comparison Across Baselines and Proposed Methods.}
    \label{fig:plot2_grounding_score}
\end{figure}

We evaluate the grounding scores of our FilterRAG framework against baseline models to assess its ability to mitigate hallucinations by aligning answers with external knowledge. As shown in Figure~\ref{fig:plot2_grounding_score}, Base1 achieves the highest grounding score of 94.60\% on the VQA v2 dataset~\cite{goyal2017making}, indicating that BLIP performs effectively when answering general-domain questions that do not require external knowledge. In contrast, Base2, evaluated on the OK-VQA dataset~\cite{marino2019ok}, shows a significant drop to 71.70\%, highlighting the challenge of answering knowledge-based questions without access to external information, thereby increasing the likelihood of hallucinations.

To address this limitation, our proposed method integrates BLIP-VQA, RAG, and external knowledge sources such as Wikipedia and DBpedia. The grounding scores for our method are 70.06\% for In-Distribution (ID) data, 70.68\% for Out-of-Distribution (OOD) data, and 70.37\% when combining both settings. These consistent scores demonstrate that FilterRAG effectively grounds answers in retrieved knowledge, reducing hallucinations even in challenging OOD scenarios.

Although our method does not achieve the grounding performance of Base1, it provides reliable results for knowledge-intensive tasks by leveraging external knowledge sources. This makes FilterRAG a robust and efficient solution for real-world VQA applications, particularly where external knowledge and OOD generalization are critical.

\subsection{Ablation Study}
\begin{figure}[h!]
    \centering
    \includegraphics[width=\linewidth]{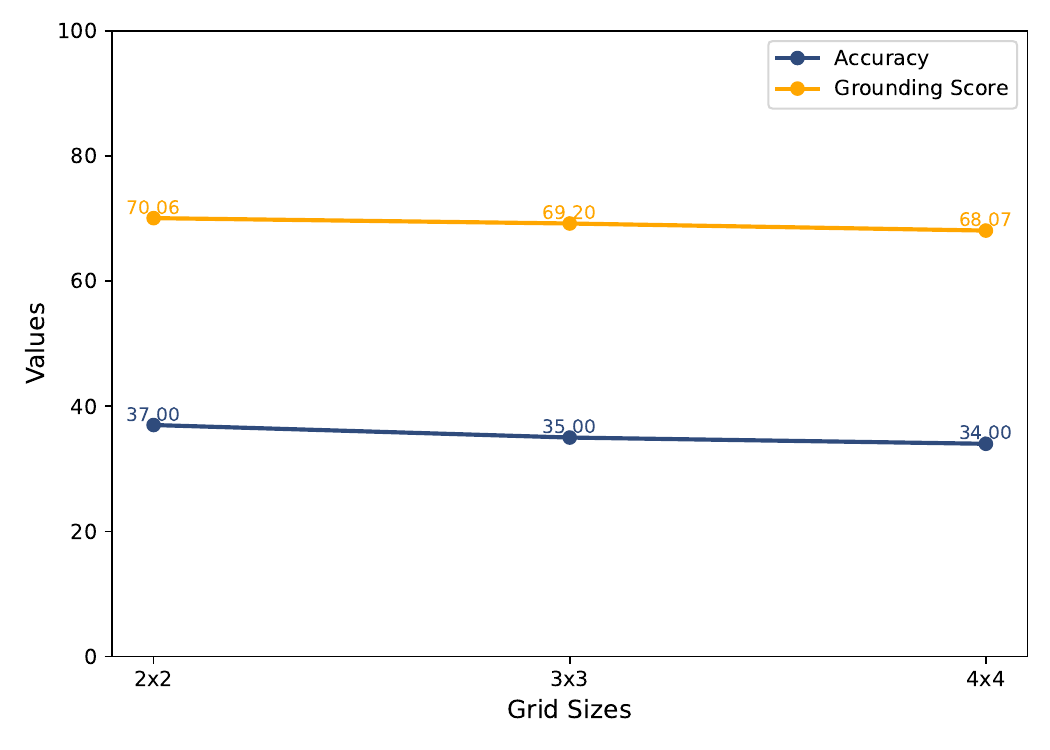}
    \caption{Effect of Grid Sizes on Accuracy and Grounding Score.}
    \label{fig:plot5_measure_grid_size}
\end{figure}

We evaluate the effect of different image grid sizes on the performance of our FilterRAG framework with BLIP-VQA and RAG in OOD scenarios. We consider three grid configurations, 2x2, 3x3, and 4x4, and evaluate their influence on accuracy and grounding score. As shown in Figure~\ref{fig:plot5_measure_grid_size}, accuracy decreases slightly as the grid size increases. The accuracy is 37.00\% for the 2x2 grid, declines to 35.00\% for the 3x3 grid, and further drops to 34.00\% for the 4x4 grid. This downward trend indicates that larger grid sizes lead to excessive fragmentation, making it challenging for the model to extract coherent and meaningful visual features.

Similarly, the grounding score follows a declining trend with increasing grid size. The grounding score is 70.06\% for the 2x2 grid, reducing to 69.20\% for the 3x3 grid and 68.07\% for the 4x4 grid. This decline suggests that finer grid divisions hinder the model’s ability to align generated answers with retrieved external knowledge, likely due to the loss of contextual coherence when images are broken into smaller patches.

Overall, the 2x2 grid size achieves the best trade-off between accuracy and grounding score. It maintains both visual coherence and effective knowledge alignment, thereby reducing the risk of hallucinations. Consequently, for OOD scenarios in the FilterRAG framework, the 2x2 grid configuration is the most effective for ensuring robust and reliable performance.

\subsection{Qualitative Analysis}
We perform a qualitative analysis of FilterRAG on the OK-VQA dataset~\cite{marino2019ok}, evaluating its performance in both In-Domain (ID) and Out-of-Distribution (OOD) settings. As illustrated in Figure~\ref{fig:Qualitative_Analysis}, FilterRAG generates accurate answers in ID scenarios where the retrieved knowledge is relevant and aligns well with the visual context. In these cases, the model effectively combines visual cues and external knowledge, resulting in well-grounded responses. These errors are frequently caused by misalignment between the visual context and the retrieved information, reflecting the challenge of handling ambiguous or novel queries outside the training distribution.

In OOD settings, FilterRAG struggles when relevant knowledge of unfamiliar concepts cannot be effectively retrieved. This often leads to hallucinations, where the model produces plausible but incorrect answers that are not supported by the retrieved evidence. This analysis highlights the critical role of reliable knowledge retrieval and precise multimodal alignment in mitigating hallucinations. Improving the quality of knowledge retrieval and refining visual-textual alignment are essential steps toward making FilterRAG more reliable in OOD contexts. Future improvements in these areas can help ensure more accurate and context-aware responses in real-world VQA applications.

%% file: sections/5_Conclusion.tex
\section{Conclusion}
\label{sec:Conclusion}
We introduced FilterRAG, a framework combining BLIP-VQA with Retrieval-Augmented Generation (RAG) to reduce hallucinations in Visual Question Answering (VQA), particularly in out-of-distribution (OOD) scenarios. By grounding responses in external knowledge sources like Wikipedia and DBpedia, FilterRAG improves accuracy and robustness for knowledge-intensive tasks. Evaluations on the OK-VQA dataset show an accuracy of 36.5\%, demonstrating its effectiveness in handling both in-domain and OOD queries. This work underscores the importance of integrating external knowledge to enhance VQA reliability. Future work will focus on improving knowledge retrieval and multimodal alignment to further reduce hallucinations and enhance generalization.

\section{Acknowledgements}
Author Sarwar gratefully acknowledges the Department of Computer Science at the University of Maryland Baltimore County (UMBC) for providing financial support through a Graduate Assistantship.

%% file: supplement.tex

\clearpage
\begin{figure}[!h]
    \centering
    \begin{minipage}{\textwidth}
        \centering
        \includegraphics[width=0.8\textwidth]{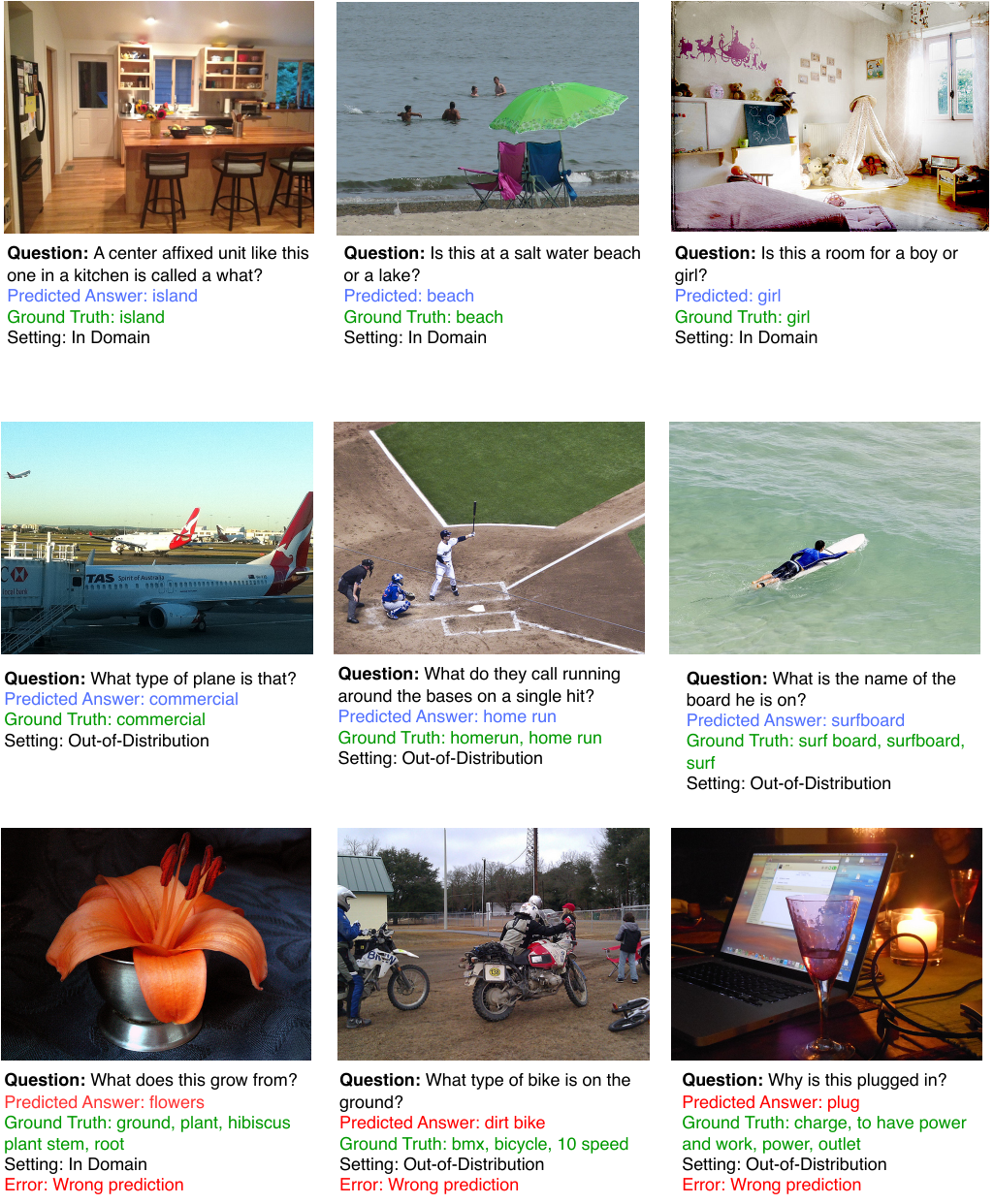}
        \captionsetup{justification=raggedright, singlelinecheck=false}  
        \caption{Qualitative Analysis of FilterRAG Predictions on OK-VQA in in-distribution (ID) and out-of-distribution (OOD) Settings. The figure illustrates the performance differences between ID and OOD settings, highlighting key areas where the model excels or fails.}
        \label{fig:Qualitative_Analysis}
    \end{minipage}
\end{figure}

%% file: main.bib
@String(AAAI = {AAAI})

@article{tu2014joint,
  title={Joint video and text parsing for understanding events and answering queries},
  author={Tu, Kewei and Meng, Meng and Lee, Mun Wai and Choe, Tae Eun and Zhu, Song-Chun},
  journal={IEEE MultiMedia},
  volume={21},
  number={2},
  pages={42--70},
  year={2014},
  publisher={IEEE}
}

@inproceedings{yang2022empirical,
  title={An empirical study of gpt-3 for few-shot knowledge-based vqa},
  author={Yang, Zhengyuan and Gan, Zhe and Wang, Jianfeng and Hu, Xiaowei and Lu, Yumao and Liu, Zicheng and Wang, Lijuan},
  booktitle={Proceedings of the AAAI conference on artificial intelligence},
  volume={36},
  number={3},
  pages={3081--3089},
  year={2022}
}

@article{shevchenko2021reasoning,
  title={Reasoning over vision and language: Exploring the benefits of supplemental knowledge},
  author={Shevchenko, Violetta and Teney, Damien and Dick, Anthony and Hengel, Anton van den},
  journal={arXiv preprint arXiv:2101.06013},
  year={2021}
}

@article{luo2021weakly,
  title={Weakly-supervised visual-retriever-reader for knowledge-based question answering},
  author={Luo, Man and Zeng, Yankai and Banerjee, Pratyay and Baral, Chitta},
  journal={arXiv preprint arXiv:2109.04014},
  year={2021}
}

@article{zhu2020mucko,
  title={Mucko: Multi-layer cross-modal knowledge reasoning for fact-based visual question answering},
  author={Zhu, Zihao and Yu, Jing and Wang, Yujing and Sun, Yajing and Hu, Yue and Wu, Qi},
  journal={arXiv preprint arXiv:2006.09073},
  year={2020}
}

@inproceedings{li2020boosting,
  title={Boosting visual question answering with context-aware knowledge aggregation},
  author={Li, Guohao and Wang, Xin and Zhu, Wenwu},
  booktitle={Proceedings of the 28th ACM International Conference on Multimedia},
  pages={1227--1235},
  year={2020}
}

@article{mokady2021clipcap,
  title={Clipcap: Clip prefix for image captioning},
  author={Mokady, Ron and Hertz, Amir and Bermano, Amit H},
  journal={arXiv preprint arXiv:2111.09734},
  year={2021}
}

@article{maynez2020faithfulness,
  title={On faithfulness and factuality in abstractive summarization},
  author={Maynez, Joshua and Narayan, Shashi and Bohnet, Bernd and McDonald, Ryan},
  journal={arXiv preprint arXiv:2005.00661},
  year={2020}
}

@article{zhou2020detecting,
  title={Detecting hallucinated content in conditional neural sequence generation},
  author={Zhou, Chunting and Neubig, Graham and Gu, Jiatao and Diab, Mona and Guzman, Paco and Zettlemoyer, Luke and Ghazvininejad, Marjan},
  journal={arXiv preprint arXiv:2011.02593},
  year={2020}
}

@article{reimers2019sentence,
  title={Sentence-BERT: Sentence Embeddings using Siamese BERT-Networks},
  author={Reimers, N},
  journal={arXiv preprint arXiv:1908.10084},
  year={2019}
}

@article{devlin2018bert,
  title={Bert: Pre-training of deep bidirectional transformers for language understanding},
  author={Devlin, Jacob},
  journal={arXiv preprint arXiv:1810.04805},
  year={2018}
}

@inproceedings{pennington2014glove,
  title={Glove: Global vectors for word representation},
  author={Pennington, Jeffrey and Socher, Richard and Manning, Christopher D},
  booktitle={Proceedings of the 2014 conference on empirical methods in natural language processing (EMNLP)},
  pages={1532--1543},
  year={2014}
}

@article{mikolov2013efficient,
  title={Efficient estimation of word representations in vector space},
  author={Mikolov, Tomas},
  journal={arXiv preprint arXiv:1301.3781},
  volume={3781},
  year={2013}
}

@article{izacard2020leveraging,
  title={Leveraging passage retrieval with generative models for open domain question answering},
  author={Izacard, Gautier and Grave, Edouard},
  journal={arXiv preprint arXiv:2007.01282},
  year={2020}
}

@article{karpukhin2020dense,
  title={Dense passage retrieval for open-domain question answering},
  author={Karpukhin, Vladimir and O{\u{g}}uz, Barlas and Min, Sewon and Lewis, Patrick and Wu, Ledell and Edunov, Sergey and Chen, Danqi and Yih, Wen-tau},
  journal={arXiv preprint arXiv:2004.04906},
  year={2020}
}

@software{gpt-neo,
  author       = {Black, Sid and
                  Leo, Gao and
                  Wang, Phil and
                  Leahy, Connor and
                  Biderman, Stella},
  title        = {{GPT-Neo: Large Scale Autoregressive Language 
                   Modeling with Mesh-Tensorflow}},
  month        = mar,
  year         = 2021,
  note         = {{If you use this software, please cite it using 
                   these metadata.}},
  publisher    = {Zenodo},
  version      = {1.0},
  doi          = {10.5281/zenodo.5297715},
  url          = {https://doi.org/10.5281/zenodo.5297715}
}

@article{sanh2021multitask,
  title={Multitask prompted training enables zero-shot task generalization},
  author={Sanh, Victor and Webson, Albert and Raffel, Colin and Bach, Stephen H and Sutawika, Lintang and Alyafeai, Zaid and Chaffin, Antoine and Stiegler, Arnaud and Scao, Teven Le and Raja, Arun and others},
  journal={arXiv preprint arXiv:2110.08207},
  year={2021}
}

@article{alayrac2022flamingo,
  title={Flamingo: a visual language model for few-shot learning},
  author={Alayrac, Jean-Baptiste and Donahue, Jeff and Luc, Pauline and Miech, Antoine and Barr, Iain and Hasson, Yana and Lenc, Karel and Mensch, Arthur and Millican, Katherine and Reynolds, Malcolm and others},
  journal={Advances in neural information processing systems},
  volume={35},
  pages={23716--23736},
  year={2022}
}

@article{tsimpoukelli2021multimodal,
  title={Multimodal few-shot learning with frozen language models},
  author={Tsimpoukelli, Maria and Menick, Jacob L and Cabi, Serkan and Eslami, SM and Vinyals, Oriol and Hill, Felix},
  journal={Advances in Neural Information Processing Systems},
  volume={34},
  pages={200--212},
  year={2021}
}

@article{wang2019survey,
  title={A survey of zero-shot learning: Settings, methods, and applications},
  author={Wang, Wei and Zheng, Vincent W and Yu, Han and Miao, Chunyan},
  journal={ACM Transactions on Intelligent Systems and Technology (TIST)},
  volume={10},
  number={2},
  pages={1--37},
  year={2019},
  publisher={ACM New York, NY, USA}
}

@article{xian2018zero,
  title={Zero-shot learning—a comprehensive evaluation of the good, the bad and the ugly},
  author={Xian, Yongqin and Lampert, Christoph H and Schiele, Bernt and Akata, Zeynep},
  journal={IEEE transactions on pattern analysis and machine intelligence},
  volume={41},
  number={9},
  pages={2251--2265},
  year={2018},
  publisher={IEEE}
}

@article{lin2022revive,
  title={Revive: Regional visual representation matters in knowledge-based visual question answering},
  author={Lin, Yuanze and Xie, Yujia and Chen, Dongdong and Xu, Yichong and Zhu, Chenguang and Yuan, Lu},
  journal={Advances in Neural Information Processing Systems},
  volume={35},
  pages={10560--10571},
  year={2022}
}

@article{gao2022thousand,
  title={A thousand words are worth more than a picture: Natural language-centric outside-knowledge visual question answering},
  author={Gao, Feng and Ping, Qing and Thattai, Govind and Reganti, Aishwarya and Wu, Ying Nian and Natarajan, Prem},
  journal={arXiv preprint arXiv:2201.05299},
  year={2022}
}

@article{fu2023generate,
  title={Generate then select: Open-ended visual question answering guided by world knowledge},
  author={Fu, Xingyu and Zhang, Sheng and Kwon, Gukyeong and Perera, Pramuditha and Zhu, Henghui and Zhang, Yuhao and Li, Alexander Hanbo and Wang, William Yang and Wang, Zhiguo and Castelli, Vittorio and others},
  journal={arXiv preprint arXiv:2305.18842},
  year={2023}
}

@article{wu2022entity,
  title={Entity-focused dense passage retrieval for outside-knowledge visual question answering},
  author={Wu, Jialin and Mooney, Raymond J},
  journal={arXiv preprint arXiv:2210.10176},
  year={2022}
}

@inproceedings{garderes2020conceptbert,
  title={Conceptbert: Concept-aware representation for visual question answering},
  author={Gard{\`e}res, Fran{\c{c}}ois and Ziaeefard, Maryam and Abeloos, Baptiste and Lecue, Freddy},
  booktitle={Findings of the Association for Computational Linguistics: EMNLP 2020},
  pages={489--498},
  year={2020}
}

@inproceedings{marino2021krisp,
  title={Krisp: Integrating implicit and symbolic knowledge for open-domain knowledge-based vqa},
  author={Marino, Kenneth and Chen, Xinlei and Parikh, Devi and Gupta, Abhinav and Rohrbach, Marcus},
  booktitle={Proceedings of the IEEE/CVF Conference on Computer Vision and Pattern Recognition},
  pages={14111--14121},
  year={2021}
}

@inproceedings{wu2022multi,
  title={Multi-modal answer validation for knowledge-based vqa},
  author={Wu, Jialin and Lu, Jiasen and Sabharwal, Ashish and Mottaghi, Roozbeh},
  booktitle={Proceedings of the AAAI conference on artificial intelligence},
  volume={36},
  number={3},
  pages={2712--2721},
  year={2022}
}

@article{gui2021kat,
  title={Kat: A knowledge augmented transformer for vision-and-language},
  author={Gui, Liangke and Wang, Borui and Huang, Qiuyuan and Hauptmann, Alex and Bisk, Yonatan and Gao, Jianfeng},
  journal={arXiv preprint arXiv:2112.08614},
  year={2021}
}

@article{guo2022survey,
  title={A survey on automated fact-checking},
  author={Guo, Zhijiang and Schlichtkrull, Michael and Vlachos, Andreas},
  journal={Transactions of the Association for Computational Linguistics},
  volume={10},
  pages={178--206},
  year={2022},
  publisher={MIT Press One Rogers Street, Cambridge, MA 02142-1209, USA journals-info~…}
}

@article{ram2023context,
  title={In-context retrieval-augmented language models},
  author={Ram, Ori and Levine, Yoav and Dalmedigos, Itay and Muhlgay, Dor and Shashua, Amnon and Leyton-Brown, Kevin and Shoham, Yoav},
  journal={Transactions of the Association for Computational Linguistics},
  volume={11},
  pages={1316--1331},
  year={2023},
  publisher={MIT Press One Broadway, 12th Floor, Cambridge, Massachusetts 02142, USA~…}
}

@inproceedings{kafle2017analysis,
  title={An analysis of visual question answering algorithms},
  author={Kafle, Kushal and Kanan, Christopher},
  booktitle={Proceedings of the IEEE international conference on computer vision},
  pages={1965--1973},
  year={2017}
}

@inproceedings{bigham2010vizwiz,
  title={Vizwiz: nearly real-time answers to visual questions},
  author={Bigham, Jeffrey P and Jayant, Chandrika and Ji, Hanjie and Little, Greg and Miller, Andrew and Miller, Robert C and Miller, Robin and Tatarowicz, Aubrey and White, Brandyn and White, Samual and others},
  booktitle={Proceedings of the 23nd annual ACM symposium on User interface software and technology},
  pages={333--342},
  year={2010}
}

@article{su2019vl,
  title={Vl-bert: Pre-training of generic visual-linguistic representations},
  author={Su, Weijie and Zhu, Xizhou and Cao, Yue and Li, Bin and Lu, Lewei and Wei, Furu and Dai, Jifeng},
  journal={arXiv preprint arXiv:1908.08530},
  year={2019}
}

@article{li2019visualbert,
  title={Visualbert: A simple and performant baseline for vision and language},
  author={Li, Liunian Harold and Yatskar, Mark and Yin, Da and Hsieh, Cho-Jui and Chang, Kai-Wei},
  journal={arXiv preprint arXiv:1908.03557},
  year={2019}
}

@inproceedings{marino2019ok,
  title={Ok-vqa: A visual question answering benchmark requiring external knowledge},
  author={Marino, Kenneth and Rastegari, Mohammad and Farhadi, Ali and Mottaghi, Roozbeh},
  booktitle={Proceedings of the IEEE/cvf conference on computer vision and pattern recognition},
  pages={3195--3204},
  year={2019}
}

@inproceedings{goyal2017making,
  title={Making the v in vqa matter: Elevating the role of image understanding in visual question answering},
  author={Goyal, Yash and Khot, Tejas and Summers-Stay, Douglas and Batra, Dhruv and Parikh, Devi},
  booktitle={Proceedings of the IEEE conference on computer vision and pattern recognition},
  pages={6904--6913},
  year={2017}
}

@inproceedings{zhang2016yin,
  title={Yin and yang: Balancing and answering binary visual questions},
  author={Zhang, Peng and Goyal, Yash and Summers-Stay, Douglas and Batra, Dhruv and Parikh, Devi},
  booktitle={Proceedings of the IEEE conference on computer vision and pattern recognition},
  pages={5014--5022},
  year={2016}
}

@article{covert2024locality,
  title={Locality Alignment Improves Vision-Language Models},
  author={Covert, Ian and Sun, Tony and Zou, James and Hashimoto, Tatsunori},
  journal={arXiv preprint arXiv:2410.11087},
  year={2024}
}

@article{li2024lamp,
  title={LaMP: Language-Motion Pretraining for Motion Generation, Retrieval, and Captioning},
  author={Li, Zhe and Yuan, Weihao and He, Yisheng and Qiu, Lingteng and Zhu, Shenhao and Gu, Xiaodong and Shen, Weichao and Dong, Yuan and Dong, Zilong and Yang, Laurence T},
  journal={arXiv preprint arXiv:2410.07093},
  year={2024}
}

@article{luo2023end,
  title={End-to-end knowledge retrieval with multi-modal queries},
  author={Luo, Man and Fang, Zhiyuan and Gokhale, Tejas and Yang, Yezhou and Baral, Chitta},
  journal={arXiv preprint arXiv:2306.00424},
  year={2023}
}

@inproceedings{li2024configure,
  title={How to configure good in-context sequence for visual question answering},
  author={Li, Li and Peng, Jiawei and Chen, Huiyi and Gao, Chongyang and Yang, Xu},
  booktitle={Proceedings of the IEEE/CVF Conference on Computer Vision and Pattern Recognition},
  pages={26710--26720},
  year={2024}
}

@article{bordes2024introduction,
  title={An introduction to vision-language modeling},
  author={Bordes, Florian and Pang, Richard Yuanzhe and Ajay, Anurag and Li, Alexander C and Bardes, Adrien and Petryk, Suzanne and Ma{\~n}as, Oscar and Lin, Zhiqiu and Mahmoud, Anas and Jayaraman, Bargav and others},
  journal={arXiv preprint arXiv:2405.17247},
  year={2024}
}

@article{zang2024overcoming,
  title={Overcoming the Pitfalls of Vision-Language Model Finetuning for OOD Generalization},
  author={Zang, Yuhang and Goh, Hanlin and Susskind, Josh and Huang, Chen},
  journal={arXiv preprint arXiv:2401.15914},
  year={2024}
}

@article{duong2023general,
  title={General-Purpose Multi-Modal OOD Detection Framework},
  author={Duong, Viet and Wu, Qiong and Zhou, Zhengyi and Zavesky, Eric and Chen, Jiahe and Liu, Xiangzhou and Hsu, Wen-Ling and Shao, Huajie},
  journal={arXiv preprint arXiv:2307.13069},
  year={2023}
}

@article{jiang2024negative,
  title={Negative label guided ood detection with pretrained vision-language models},
  author={Jiang, Xue and Liu, Feng and Fang, Zhen and Chen, Hong and Liu, Tongliang and Zheng, Feng and Han, Bo},
  journal={arXiv preprint arXiv:2403.20078},
  year={2024}
}

@article{dong2024multiood,
  title={MultiOOD: Scaling Out-of-Distribution Detection for Multiple Modalities},
  author={Dong, Hao and Zhao, Yue and Chatzi, Eleni and Fink, Olga},
  journal={arXiv preprint arXiv:2405.17419},
  year={2024}
}

@article{lu2019vilbert,
  title={Vilbert: Pretraining task-agnostic visiolinguistic representations for vision-and-language tasks},
  author={Lu, Jiasen and Batra, Dhruv and Parikh, Devi and Lee, Stefan},
  journal={Advances in neural information processing systems},
  volume={32},
  year={2019}
}

@article{tan2019lxmert,
  title={Lxmert: Learning cross-modality encoder representations from transformers},
  author={Tan, Hao and Bansal, Mohit},
  journal={arXiv preprint arXiv:1908.07490},
  year={2019}
}

@article{hendrycks2016baseline,
  title={A baseline for detecting misclassified and out-of-distribution examples in neural networks},
  author={Hendrycks, Dan and Gimpel, Kevin},
  journal={arXiv preprint arXiv:1610.02136},
  year={2016}
}

@inproceedings{singh2022flava,
  title={Flava: A foundational language and vision alignment model},
  author={Singh, Amanpreet and Hu, Ronghang and Goswami, Vedanuj and Couairon, Guillaume and Galuba, Wojciech and Rohrbach, Marcus and Kiela, Douwe},
  booktitle={Proceedings of the IEEE/CVF Conference on Computer Vision and Pattern Recognition},
  pages={15638--15650},
  year={2022}
}

@inproceedings{lin2014microsoft,
  title={Microsoft coco: Common objects in context},
  author={Lin, Tsung-Yi and Maire, Michael and Belongie, Serge and Hays, James and Perona, Pietro and Ramanan, Deva and Doll{\'a}r, Piotr and Zitnick, C Lawrence},
  booktitle={Computer Vision--ECCV 2014: 13th European Conference, Zurich, Switzerland, September 6-12, 2014, Proceedings, Part V 13},
  pages={740--755},
  year={2014},
  organization={Springer}
}

@inproceedings{jia2021scaling,
  title={Scaling up visual and vision-language representation learning with noisy text supervision},
  author={Jia, Chao and Yang, Yinfei and Xia, Ye and Chen, Yi-Ting and Parekh, Zarana and Pham, Hieu and Le, Quoc and Sung, Yun-Hsuan and Li, Zhen and Duerig, Tom},
  booktitle={International conference on machine learning},
  pages={4904--4916},
  year={2021},
  organization={PMLR}
}

@inproceedings{radford2021learning,
  title={Learning transferable visual models from natural language supervision},
  author={Radford, Alec and Kim, Jong Wook and Hallacy, Chris and Ramesh, Aditya and Goh, Gabriel and Agarwal, Sandhini and Sastry, Girish and Askell, Amanda and Mishkin, Pamela and Clark, Jack and others},
  booktitle={International conference on machine learning},
  pages={8748--8763},
  year={2021},
  organization={PMLR}
}

@article{lewis2020retrieval,
  title={Retrieval-augmented generation for knowledge-intensive nlp tasks},
  author={Lewis, Patrick and Perez, Ethan and Piktus, Aleksandra and Petroni, Fabio and Karpukhin, Vladimir and Goyal, Naman and K{\"u}ttler, Heinrich and Lewis, Mike and Yih, Wen-tau and Rockt{\"a}schel, Tim and others},
  journal={Advances in Neural Information Processing Systems},
  volume={33},
  pages={9459--9474},
  year={2020}
}

@inproceedings{li2022blip,
  title={Blip: Bootstrapping language-image pre-training for unified vision-language understanding and generation},
  author={Li, Junnan and Li, Dongxu and Xiong, Caiming and Hoi, Steven},
  booktitle={International conference on machine learning},
  pages={12888--12900},
  year={2022},
  organization={PMLR}
}

@article{brown2020language,
  title={Language models are few-shot learners},
  author={Brown, Tom and Mann, Benjamin and Ryder, Nick and Subbiah, Melanie and Kaplan, Jared D and Dhariwal, Prafulla and Neelakantan, Arvind and Shyam, Pranav and Sastry, Girish and Askell, Amanda and others},
  journal={Advances in neural information processing systems},
  volume={33},
  pages={1877--1901},
  year={2020}
}

@inproceedings{antol2015vqa,
  title={Vqa: Visual question answering},
  author={Antol, Stanislaw and Agrawal, Aishwarya and Lu, Jiasen and Mitchell, Margaret and Batra, Dhruv and Zitnick, C Lawrence and Parikh, Devi},
  booktitle={Proceedings of the IEEE international conference on computer vision},
  pages={2425--2433},
  year={2015}
}
